\documentclass[sigconf, screen, anonymous=false]{acmart}
\AtBeginDocument{%
  }

\usepackage{microtype}
\usepackage{hyperref}
\usepackage{url}
\usepackage[utf8]{inputenc} 
\usepackage[T1]{fontenc}    
\usepackage{booktabs}
\usepackage{nicefrac}       
\usepackage{microtype}      
\usepackage{xcolor}         
\usepackage{amsmath}
\usepackage{natbib}
\usepackage{caption}
\usepackage{subcaption}
\usepackage{latexsym}
\usepackage{bbding}
\usepackage{caption}
\usepackage{subcaption}
\usepackage{pifont}
\usepackage{wasysym}
\usepackage{listings}
\usepackage{graphicx}
\usepackage{makecell}
\usepackage{threeparttable}
\usepackage{wrapfig}
\usepackage{multirow}
\usepackage{colortbl}
\usepackage{enumitem}
\usepackage{algpseudocode}
\usepackage{tcolorbox}
\usepackage{afterpage}
\usepackage[linesnumbered,ruled,vlined]{algorithm2e}
\usepackage{tcolorbox}
\newtcolorbox{promptbox}{
  colback=gray!5!white, colframe=gray!60!black, boxrule=0.5pt, arc=3pt}

\newcommand{\OURS}{\text{LANTERN}\xspace}

\begin{document}









\author{
Zhoutong Fu\textsuperscript{*}\quad Yihan Cao\textsuperscript{*}\quad Yi-Lin Chen\quad Aman Lunia\quad Liming Dong\quad Neha Saraf\quad Ruijie Jiang\quad Yun Dai\textsuperscript{\dag}\quad Qingquan Song\quad Tan Wang\quad Guoyao Li\quad Derek Koh\quad Haichao Wei\textsuperscript{\dag}\quad Zhipeng Wang\quad Aman Gupta\textsuperscript{\dag}\quad Chengming Jiang\quad Jianqiang Shen\quad Liangjie Hong\quad Wenjing Zhang \\
\texttt{\{zfu, yihacao, yilchen, alunia, mdong, nsaraf, rjiang, yudai, qsong, twang2, guoyli, dkoh, hawei, zhipwang, amagupta, cjiang, jershen, liahong, wzhang\}@linkedin.com} \\
LinkedIn, Mountain View, CA, USA
}

\title{LANTERN: Scalable Distillation of Large Language Models for Job-Person Fit and Explanation} 

\renewcommand{\shortauthors}{Trovato et al.}

\begin{abstract}

Large language models (LLMs) have achieved strong performance across a wide range of natural language processing tasks. However, deploying LLMs at scale for domain-specific applications—such as job-person fit and explanation in job-seeking platforms—introduces distinct challenges. At LinkedIn, the job–person fit task requires analyzing a candidate’s public profile against job requirements to produce both a fit assessment and a detailed explanation. Directly applying open-source or finetuned LLMs to this task often fails to yield high-quality, actionable feedback due to the complexity of the domain and the need for structured outputs. Moreover, the large size of these models leads to high inference latency and limits scalability, making them unsuitable for online use. To address these challenges, we introduce LANTERN, a novel LLM knowledge distillation framework tailored specifically for job-person fit tasks. LANTERN involves modeling over multiple objectives, an encoder model for classification purpose, and a decoder model for explanation purpose. To better distill the knowledge from a strong black-box teacher model to multiple downstream models, LANTERN incorporates multi-level knowledge distillation that integrates both data and logit-level insights. In addition to introducing the knowledge distillation framework, we share our insights on post-training techniques and prompt engineering, both of which are crucial for successfully adapting LLMs to domain-specific downstream tasks. Extensive experimental results demonstrate that LANTERN significantly improves task-specific metrics for both job–person fit and explanation. Online evaluations further confirm its effectiveness, showing measurable gains in job seeker engagement, including +0.24\% increase in apply rate and a +0.28\% increase in qualified applications.
\end{abstract}




\maketitle
\renewcommand{\thefootnote}{\fnsymbol{footnote}}
\footnotetext[1]{Equal contribution.}
\footnotetext[2]{Work done while at LinkedIn.}
\section{Introduction}
Recent advances in large language models (LLMs) have transformed natural language processing (NLP)~\cite{minaee2024large, lai2025survey,cao2023comprehensive}, enabling machines to generate coherent, context-aware, and human-like outputs~\cite{tie2025large, ding2023enhancing}. In recommendation systems and personalized applications, LLMs offer new capabilities for understanding user preferences, generating tailored content, and enhancing user engagement~\cite{lyu2023llm,wu2024survey}.
At LinkedIn, we harness the powerful understanding capabilities of LLMs to deliver precise recommendations and insights for both job seekers and hirers~\cite{borisyuk2024lirank,borisyuk2024lignn}. In this paper, we present a detailed overview of our training and serving system designed to enhance job-person fit prediction and explanation through LLM distillation

\begin{figure}[t]
    \centering
    \includegraphics[width=0.85\linewidth]{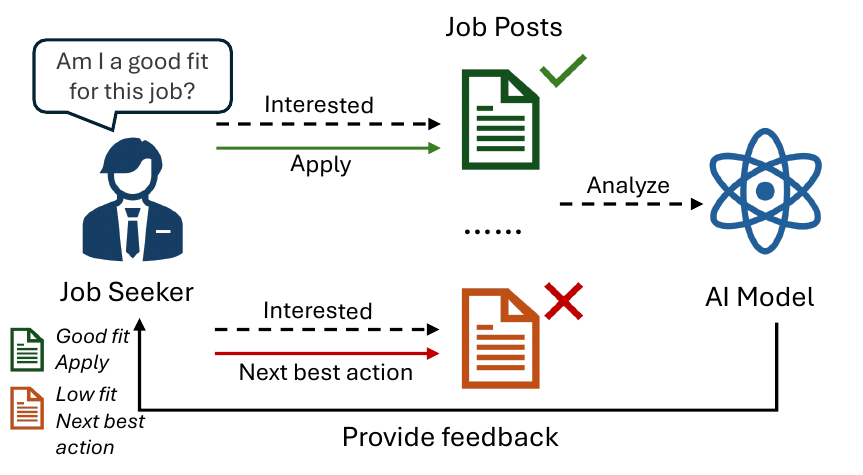}
    \caption{The illustration of how the AI model assists job seekers by evaluating job–person fit. When a user expresses interest in a job, the model analyzes the posting and recommends one of two actions: apply if the fit is high, or pursue a next-best alternative if the fit is low. User behavior, such as whether they apply or not, provides implicit feedback, which is used to further refine the model’s recommendations over time.}
    \label{fig:task}
\end{figure}

In the job marketplace, both job seekers and hirers aim to find optimal matches. Job seekers seek relevant, achievable opportunities, while hirers aim to identify qualified candidates. The job-person fit task supports both goals by evaluating how well a candidate’s experience aligns with a job’s requirements, and providing interpretable feedback~\cite{qin2020enhanced,cao2024tarot}. In our product use case, given a candidate profile and a job description, the system produces a fit level (e.g., Low, Medium, High) and a line-by-line explanation outlining how the qualifications are met, as shown in Figure~\ref{fig:task}.


Traditional approaches~\cite{qin2020enhanced,jiang2020learning} to this task rely heavily on manual evaluation or user interaction signals, both of which pose limitations. User-provided signals are often noisy, biased, and inconsistent, while manually labeled training data is expensive and slow to scale. LLMs offer an opportunity to synthesize large quantities of supervision automatically, but using them directly introduces new challenges.

\textbf{Scalability and latency}. The large size of most state-of-the-art LLMs leads to high inference latency and significant resource costs, making them impractical for online serving.

\begin{figure}[t]
    \centering
    \includegraphics[width=0.85\linewidth]{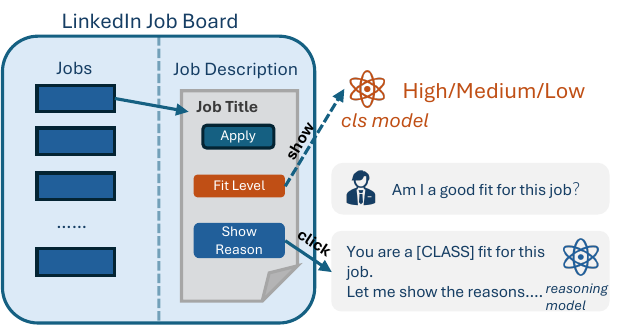}
    \caption{Illustration of how \OURS is integrated into the LinkedIn Job Board. When a user views a job description, the classification model predicts a fit label (High/Medium/Low), which is shown alongside the job. Upon clicking “Show Reason,” the explanation model generates a detailed, personalized rationale, helping the user understand why they are a good match for the job.}
    \label{fig:job-board}
\end{figure}

\textbf{Accuracy and interpretability}. The job-person fit task requires dual outputs: (1) a classification of fit and (2) a detailed explanation, as shown in Figure~\ref{fig:job-board}. This complexity makes it difficult for general-purpose LLMs to produce accurate, structured outputs with minimal finetuning. These challenges call for a more targeted and efficient solution.

To address these challenges, we introduce \OURS, a novel distillation-based training framework that transfers knowledge from a powerful black-box teacher LLM into two lightweight student models: an encoder-based classification model and a decoder-based explanation model, which are more suitable for online serving, as shown in figure~\ref{fig:structure}. 

\OURS employs both black-box and white-box knowledge distillation, enabling the student model to learn not only data-level patterns but also logit-level patterns, ensuring a comprehensive transfer of knowledge. 
In addition to introducing the distillation framework, we share useful insights on prompt engineering for the effective training of LLMs.

We evaluate \OURS from two dimensions: accuracy and latency.
Extensive experiment results show that \OURS shows both high-quality explanation and classification accuracy, demonstrating its effectiveness in transferring domain-specific knowledge for job–person fit tasks.
Online deployment demonstrates measurable impact, including a +0.24\% increase in apply rate and a +0.28\% increase in qualified applications.

We summarize our contributions as follows:
\begin{itemize}
    \item We propose \OURS, a scalable knowledge distillation framework tailored for job-person fit and explanation, which jointly distills a black-box large generative language model into a lightweight classification model and a generative explanation model.
    \item We provide novel insights into synthetic data generation and post-training strategies, emphasizing the integration of both white-box and black-box distillation approaches to effectively transfer knowledge from large teacher models to compact student models.
    \item We conduct extensive offline and online evaluations, showing that \OURS achieves high-quality outputs with low latency, making it suitable for real-time applications
\end{itemize}

\section{Related Work}
\subsection{Large Language Models}
Large language models (LLMs) have recently emerged as powerful tools for evaluating job-person fit, demonstrating strong capabilities in understanding, analyzing, and comparing textual information from public profiles and job descriptions~\cite{wei2025control,cao2024tarot}. Several novel LLM-based frameworks have emerged to address job-person fit challenges. One approach focuses on leveraging LLMs for behavioral graph analysis in job recommendations, introducing meta-path prompt constructors to help LLMs understand behavioral graph semantics and path augmentation modules to reduce prompt bias from sequence-based inputs~\cite{wu2024exploring,maree2024optimizing}. This framework enables personalized job recommendations by analyzing underlying patterns and relationships in user behavior data.
The integration of LLMs with traditional person-job fit methods has expanded beyond simple text matching. Current research explores how LLMs can refine documents to handle linguistic complexity and integrate structured knowledge to improve both accuracy and interpretability~\cite{hruschkanatural}. These approaches build upon existing transformer-based methods that adapt BERT's next-sentence prediction task specifically for resume-job matching~\cite{kaya2023exploration}, while incorporating fine-grained factors such as skills~\cite{alsaif2022learning, yao2022knowledge}, experience levels~\cite{dong2021analysis}, and other contextual information~\cite{luo2019resumegan}.

\subsection{Job-Person Fit}
Job-person fit, also known as Person-job Fit (P-J Fit) under some circumstances, aims to match candidates and job posts on online recruitment platforms using machine learning algorithms~\cite{jiang2020learning}. Early approaches treated this as a recommendation problem and applied collaborative filtering algorithms, but these methods ignored the actual content of job posts and resumes, such as candidate work experience and specific job requirements~\cite{jiang2020learning}.
The field has since evolved to leverage deep learning models, which have significantly improved performance on natural language processing tasks including semantic matching~\cite{jiang2020learning}.
These deep learning-based methods focus on learning effective representations of the free text content in both job posts and resumes~\cite{bian2019domain,qin2020enhanced}.

\subsection{Knowledge Distillation}
Knowledge distillation methods for large language models can be broadly categorized into white-box and black-box approaches~\cite{zhang2024dual, liu2024ddk}. White-box KD leverages either the internal parameters or the logits of the teacher LLM during the distillation process, while black-box KD relies solely on the outputs from the teacher model.  The foundational approach uses a simple loss function that minimizes KL-Divergence between output-class probability distributions of teacher and student models ~\cite{yao2024distilling}. However, recent work has identified significant limitations with the traditional forward KL divergence objective. The forward KL approach can lead to "mode-covering" behavior~\cite{gu2023minillm},  where student models overestimate low-probability regions of the teacher's distribution, potentially causing hallucinations and low-quality outputs. Multi-granularity approaches have emerged as another important methodological advancement. These methods gather intermediate representations from multiple semantic levels including tokens, spans, and samples, forming knowledge as sophisticated structural relations based on pair-wise interactions and triplet-wise geometric angles~\cite{liu2022multi}. 

\begin{figure*}[t]
    \centering
    \includegraphics[width=0.95\linewidth]{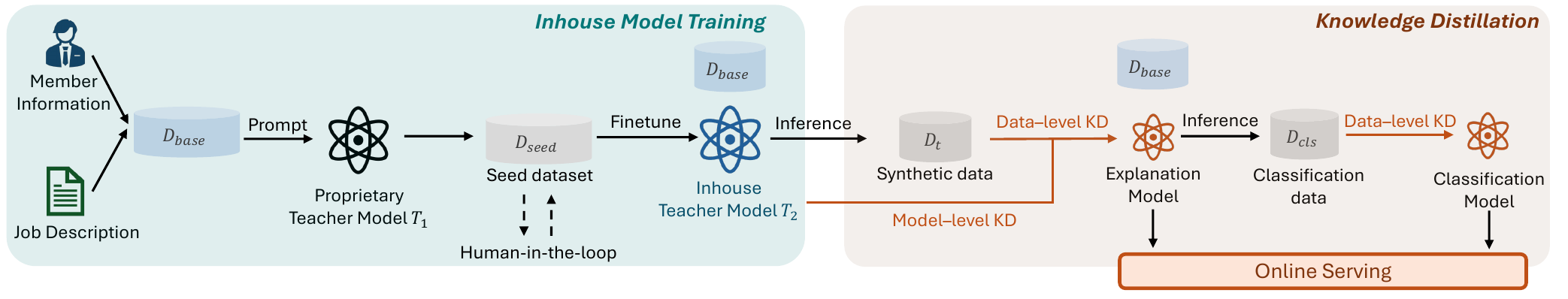}
    \caption{\OURS structure. The framework consists of two stages: in-house model training (left) and knowledge distillation (right). In the first stage, member information and job descriptions are collected to form a base dataset $D_{\text{base}}$, which is used to prompt a proprietary teacher model $T_1$. The generated outputs are curated via human-in-the-loop filtering to form a high-quality seed dataset $D_{\text{seed}}$ for finetuning an in-house teacher model $T_2$. In the second stage, $T_2$ is used to generate synthetic data $D_t$ for distilling an explanation model and corresponding classification data $D_{\text{cls}}$ for training a classification model. Both models are supervised via data-level knowledge distillation and are deployed for online serving.}
    \label{fig:structure}
\end{figure*}

\section{\OURS}
In this section, we formally define the job-person fit task, introduce \OURS, and provide a detailed explanation of its structure and implementation. The name reflects the framework’s goal of shedding light on job-person fit and explanation, addressing real-world challenges.

\subsection{Problem Setup}
Let $m$ denote a member and $j$ denote a job. Given a member-job pair $(m, j)_i$, the goal of the job-person fit task is to generate two outputs:
\begin{itemize}
    \item A match rating $r_i \in \mathcal{R}$, indicating how well member $m$ fits job $j$
    \item An explanation $exp_i \in \mathcal{E}$, providing a natural language justification for the predicted rating.
\end{itemize}
Here, $\mathcal{R}$ denotes the space of possible rating values, defined as the closed interval $[0, 1]$. A higher score corresponds to a stronger match. $\mathcal{E}$ denotes the explanation space, which consists of natural language sequences.

At LinkedIn, the match rating $r_i$ is queried 30$\times$ more often than the explanation $exp_i$, as $r_i$ is shown by default while $exp_i$ is only displayed upon explicit user request. This discrepancy poses a challenge in balancing latency and generation quality across the two outputs.

To address this, we adopt a two-model architecture:
\begin{itemize}
    \item $\text{LM}_{cls}$ predicts the score $r_i$ based on the input pair $(m, j)_i$;
    \item $\text{LM}_{exp}$ generates $exp_i$ conditioned on the same input and optionally on $r_i$.
\end{itemize}

This decoupled setup allows us to optimize each model independently according to its usage frequency and system requirements. However, designing two models in isolation also raises new challenges, such as ensuring consistency between the predicted fit levels and the generated explanation, and efficiently training both models with limited supervision. In the following sections, we introduce our approach to address these issues.


\subsection{Overall Architecture}
As illustrated in Figure~\ref{fig:structure}, our framework consists of two stages: (1) \textbf{in-house model training} and (2) \textbf{knowledge distillation}. We start from constructing a base dataset $D_{\text{base}}$ from member information and job descriptions. 
A strong black-box teacher (e.g., GPT-4o~\cite{hurst2024gpt}) is then prompted on $D_{\text{base}}$ to produce fit levels and explanatory text.
Expert annotators review these outputs and retain only high-quality responses, yielding the seed dataset $D_{\text{seed}}$. Finally we finetune an in-house teacher model on $D_\text{seed}$. 
In the next KD stage, we aim to distill the knowledge from the in-house teacher model into two separate student models, the explanation model and the classification model.

\textbf{Explanation Model.}  We perform white-box distillation on $D_{\text{seed}}$, training the student model to mimic the teacher's distribution $\pi_{T_2}$ and thus generate detailed match explanations. We specifically focus on the generation ability of this model.

\textbf{Classification Model.} We have the teacher generate labels over a large, unlabeled corpus derived from member profile with job descriptions. This labeled dataset is then used to train the classification student model to predict the match ratings.

We provide the details and design for each step in the following sections.

\subsection{In-House Teacher Model Training}
In this section, we detail our prompt design on how to get the labels from strong black-box models and use this result for in-house teacher model training.

\subsubsection{Prompt Engineering}\label{sec:prompt}
Following prompt design instructions from prior works, we separate the prompt into several structured sections to guide the model in performing job–person fit with chain-of-thought style instructions. The overall prompt structure is as follows.

\begin{itemize}
    \item \textbf{Task Instruction}: Specifies the objective of the task and defines what the language model is expected to accomplish.
    \item \textbf{Job Information Extraction Guidelines}: Provides detailed instructions for extracting relevant qualification information from the job description.
    \item \textbf{Evaluation Guidelines}: Outlines criteria for assessing whether a given user profile aligns with the extracted job qualifications.
    \item \textbf{Reasoning Guidelines}: Encourages the model to reason before making a final decision. The model is instructed to generate both extraction reasoning and evaluation reasoning, followed by a categorical score indicating the degree of fit.
    \item \textbf{Output Guidelines}: Defines constraints and formatting requirements for the model's output.
\end{itemize}

We found that prompting models, including reasoning-focused LLMs, with a long and complex instruction block often leads to inconsistent or incomplete outputs. To improve performance, we further split the prompt into a sequence of smaller subtasks (e.g., extraction followed by evaluation), reuse intermediate LLM outputs across steps, and generate a final summary that produces the succinct reasoning and desired output. This decomposition reduces prompt length and improves both output quality and consistency. Below is a pseudo example illustrating the extraction subtask; the actual prompt is omitted for confidentiality.


\begin{promptbox}
\textbf{Subtask:} Extract job qualifications. \\
\textbf{Input:} Raw job description text. \\
\textbf{Output:} A list of job qualifications with contextual descriptions.

\textbf{Extraction Guidelines:}
\begin{itemize}
    \item Identify sections labeled with terms such as “Qualifications,” “Requirements,” or similar.
    \item Extract all listed qualifications verbatim when clearly structured or infer qualifications based on keywords.
    \item Rephrase short or ambiguous phrases into full qualification statements.
    \item Preserve indicators of importance or preference.
    \item Reason about how to achieve the above steps.
\end{itemize}
\end{promptbox}

\subsubsection{Teacher Model Training}
As shown in Figure~\ref{fig:structure}, we construct the in-house teacher model by finetuning on a seed dataset $D_{\text{seed}}$, which is distilled from a proprietary black-box teacher. We begin by collecting member–job pairs that include rich profile attributes and detailed job descriptions. These inputs are formatted using the prompting templates described in Section~\ref{sec:prompt} and fed into the black-box teacher to generate both structured labels and explanatory outputs. The resulting examples form the initial candidate pool, which is further refined through human-in-the-loop filtering to retain only high-quality samples. The resulting dataset $D_{\text{seed}}$ is then used to train the in-house teacher model. We share the detailed training settings in section~\ref{sec:training-detail}.

\subsection{Knowledge Distillation}
In this section, we describe the second component of our system, corresponding to the orange module in Figure~\ref{fig:structure}. The final output artifacts consist of two models: an explanation model that generates detailed job–person analyses to justify why a candidate is a good, medium, or low fit; and a classification model that directly outputs matching results indicating the candidate's fit level for a given job.

\subsubsection{Explanation Model}

The explanation model $\text{LM}_{\text{exp}}$ is responsible for generating natural language rationales that justify the predicted match between a member and a job. To balance generation quality and inference efficiency, we adopt a knowledge distillation framework in which a larger teacher model guides the training of a smaller student model.

Specifically, we use a high-capacity teacher language model (e.g., Qwen2.5-7B) to generate high-quality explanations based on the input $x_i = (m, j)_i$ and optionally the predicted score $r_i$. The student model (e.g., Qwen2.5-1.5B or 0.5B) is then trained to mimic the teacher’s behavior while also learning from human-labeled or curated reference explanations $y_i = (y_{i,1}, \dots, y_{i,T})$ when available.

For each sample, the total training loss $\ell_{\text{exp}}$ consists of two components:

\paragraph{Supervised Fine-Tuning Loss.}
The student model $\pi_s$ is trained to maximize the likelihood of the reference explanation:
\begin{equation}
\ell_{\text{sft}} = - \sum_{t=1}^{T} \log \pi_s(y_{i,t} \mid y_{i,<t}, x_i).
\end{equation}

\paragraph{Knowledge Distillation Loss.}
To transfer knowledge from the teacher model $\pi_t$, we apply a distillation loss that aligns the teacher and student distributions at each decoding step:
\begin{equation}
\ell_{\text{kd}} = \sum_{t=1}^{T} d\left( \pi_t(\cdot \mid y_{i,<t}, x_i), \; \pi_s(\cdot \mid y_{i,<t}, x_i) \right),
\end{equation}
where $d(\cdot,\cdot)$ denotes a divergence or distance metric. We experiment with several choices for $d$, including KL divergence and Jensen–Shannon divergence, each applied independently. During training, the teacher model $\pi_t$ is frozen, and only the student model $\pi_s$ is updated. Detailed configurations are provided in the experimental section.

\paragraph{Final Objective.}
The total loss for a single example is a weighted combination of the two components:
\begin{equation}
\ell_{\text{exp}} = \lambda_{\text{sft}} \cdot \ell_{\text{sft}} + \lambda_{\text{kd}} \cdot \ell_{\text{kd}}, \label{eq:single_sample_loss}
\end{equation}
where $\lambda_{\text{sft}}$ and $\lambda_{\text{kd}}$ control the trade-off between reference supervision and teacher guidance.

\begin{figure}[t]
    \centering
    \includegraphics[width=0.75\linewidth]{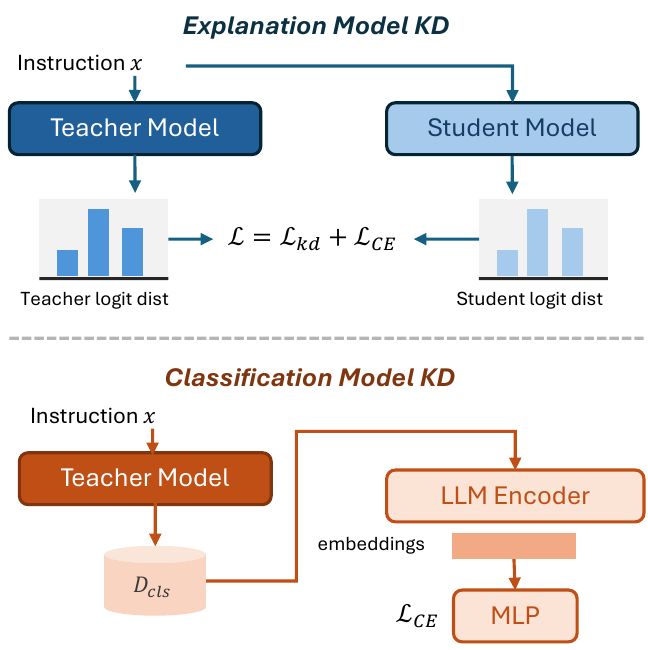}
    \caption{\OURS distillation structure.}
    \label{fig:distill}
\end{figure}

\subsubsection{Classification Model}
To distill knowledge into a classification model, we adopt a large language model (LLM) as the encoder backbone, as illustrated in Figure~\ref{fig:distill}. The input consists of synthetic job–person pairs $(x, y)$ generated by the teacher model, where $x$ is the concatenation of the job description and profile, and $y \in {1, 2, 3}$ denotes the ground-truth fit label (e.g., low, medium, or high).

The input $x$ is first encoded by the frozen or finetuned LLM encoder $f_{\text{LLM}}$ to obtain a contextualized sentence embedding:
\begin{equation}
    h=f_{\text{LLM}}(x)
\end{equation}
This embedding is then passed through a lightweight classification head $g_{\text{MLP}}$, followed by a softmax layer to produce a probability distribution over the fit categories:
\begin{equation}
    \hat{z}=softmax(g_{\text{MLP}}(h))
\end{equation}

The model is trained using a cross-entropy loss:

\begin{equation}
    \ell_{\text{cls}}=-\sum_{c=1}^C z_c \log \hat z_c
\end{equation}
where $C$ is the number of fit categories and $z_c$ is the one-hot label.

\subsection{Model Serving}
\label{sec:serving}
To meet real-time latency requirements while maintaining high throughput, we introduce two key optimizations: (1) input compression for job descriptions and (2) model size reduction for classification.

\subsubsection{Input Compression}

We compress job descriptions by extracting requirement specifications, which typically carry the strongest signal for job-person fit while other parts, such as company benefits or perks, contribute less signal. This compression is only applied to job descriptions, not to user input, such as resumes or profiles, which are more dynamic and frequently updated. Compression is triggered upon job creation and performed using a small generative model fine-tuned following the extraction template in the prompt engineering section for consistency. The resulting compressed text is stored and reused for both classification and explanation inference.

\subsubsection{Classification Model Size Reduction}

Input compression simplifies the classification task by focusing on a concise list of job requirements rather than a full-length job description. This narrower focus enables us to experiment with smaller models with similar or slight degradation performance.

\subsubsection{Online Serving Pipeline}

The final LANTERN serving pipeline consists of three modules:

\begin{itemize}
\item \textbf{Summarization Module}: Compresses the full job description into a focused summary that highlights requirement specifications during job ingestion.

\item \textbf{Classification Module}: Takes the compressed job description and user profile as input to predict a fit level (e.g. low, medium, or high). By default, a lightweight encoder model is used, with a multilingual variant used as a fallback for broader language support.

\item \textbf{Explanation Module}: Generates a line-by-line natural language rationale explaining the predicted fit levels. This module shares the same input as the classification module and produces interpretable outputs for transparency.
\end{itemize}

\begin{table*}[t]
\centering
\resizebox{0.8\textwidth}{!}{
\begin{tabular}{l|c|cccccc}
    \toprule
    \textbf{Teacher Model} & \textbf{Student Model} & \textbf{KD Method} & \textbf{Loss} &  \textbf{ROUGE\_1} & \textbf{ROUGE\_2} &  \textbf{ROUGE\_L}\\
    \midrule
    \multirow{4}{*}{Qwen2.5-7B} &
    \multirow{4}{*}{Qwen2.5-1.5B} 
        & JS & 0.1167  & 0.7820 & 0.6390 & 0.6715 \\
    &   & FKL & 0.1148  & 0.7775 & 0.6342 & 0.6670 \\
    &   & \textbf{TVD} & \textbf{0.1193}  & \textbf{0.7889}& \textbf{0.6461}& \textbf{0.6791} \\
    &   & SKL & 0.1147  &0.7858& 0.6422 & 0.6735  \\
    \bottomrule
\end{tabular}
}
\caption{Explanation task results. Loss refers to NLL loss.}
\label{table:main-results}
\end{table*}

\section{Implementation Details}
This section outlines implementation details, including data collection procedures, training configurations, and descriptions of baselines and evaluation metrics.

\subsection{Data Collection}

\subsubsection{Input Data Sampling}

The initial seed dataset is constructed by simulating pseudo job recommendations using production infrastructure, combined with stratified sampling to ensure a balanced distribution of job–profile fit categories.

For the explanation task, input samples are collected from various sources in the production tracking data. We perform downsampling on dominant sources and normalize the class for a balance distribution. 

For the classification task, we add more negative samples using user interaction signals to improve coverage of borderline or unclear matches.

\subsubsection{Data Generation}
\label{sec:data}
We generate training labels using both in-house and proprietary teacher models, and use a decoding temperature of 0 to reduce hallucinations and produce consistent, fact-based outputs.

\subsubsection{Human-in-the-Loop~\cite{wu2022survey} Evaluation}
\label{sec:human-eval}
A portion of the model-generated outputs is reviewed by human evaluators, who check for hallucinations such as incorrect information or unsupported rationale. This helps monitor training data quality and filter out low-quality datasets.

\subsection{Training Details}\label{sec:training-detail}
All training is performed on a single node with 8 NVIDIA H100 80GB GPUs, unless otherwise specified.

\paragraph{Teacher Model Training.}  
To train the in-house teacher model, we curate a high-quality dataset derived from black-box model outputs, as described in Section~\ref{sec:data}. We finetune the model for 4 epochs using the AdamW optimizer with a learning rate of $2 \times 10^{-5}$. The maximum sequence length is set to 8192, based on the distribution of input lengths in the training set. We adopt the \texttt{Qwen2.5}\footnote{https://huggingface.co/collections/Qwen/qwen25-66e81a666513e518adb90d9e}~\cite{qwen2025qwen25technicalreport} model series for both the base and finetuned teacher models, as it offers a flexible range of model sizes from 0.5B to 32B.

\paragraph{Explanation Model Distillation.}
For explanation model distillation, we leverage the same training samples used for teacher model training and distill the output distribution from the teacher model $\pi_t$ into the student model $\pi_s$. We explore several knowledge distillation objectives, including Jensen–Shannon divergence (JS), forward KL (FKL), total variation distance (TVD)~\cite{bhattacharyya2022approximating}, and symmetric KL (SKL), defined as follows:
\begin{align*}
&\text{JS}(\pi_t \| \pi_s) := \frac{1}{2} \cdot \text{KL}(\pi_t \| m) + \frac{1}{2} \cdot \text{KL}(\pi_s \| m), \text{where } m = \frac{1}{2}(\pi_t + \pi_s), \\
&\text{FKL}(\pi_t \| \pi_s) := \sum_j \pi_t(j) \log \frac{\pi_t(j)}{\pi_s(j)}, \\
&\text{TVD}(\pi_t, \pi_s) := \frac{1}{2} \sum_j \left| \pi_t(j) - \pi_s(j) \right|, \\
&\text{SKL}(\pi_t \| \pi_s) := \sum_j \pi_t(j) \log \frac{\pi_t(j)}{\pi_s(j)} 
+ \sum_j \pi_s(j) \log \frac{\pi_s(j)}{\pi_t(j)}.
\end{align*}
Each divergence provides a different lens to measure distributional discrepancy, enabling us to assess the impact of the distillation objective on student performance and convergence behavior.

By instantiating $\ell_{\text{kd}}$ with one of the divergences above and $\ell_{\text{sft}}$ with the standard cross-entropy loss, we define the explanation distillation loss as:

$$
\ell_{\text{exp}} = \lambda_{\text{sft}} \cdot \ell_{\text{CE}} + \lambda_{\text{KD}} \cdot \ell_{\text{kd}}(\pi_t \| \pi_s),
$$

where $\lambda_{\text{sft}}$ and $\lambda_{\text{KD}}$ control the trade-off between ground-truth supervision and distillation.

The final training objective is the empirical risk over all examples:

$$
\mathcal{L}_{\text{exp}} = \frac{1}{N} \sum_{i=1}^N \ell_{\text{exp}}^{(i)}.
$$

We tune $\lambda_{\text{sft}} = 0.1$ and $\lambda_{\text{KD}} = 0.9$ on a held-out validation set. Training follows the same hardware setup as the teacher model. We fine-tune the student model for 8 epochs using the AdamW optimizer with a learning rate of $8 \times 10^{-5}$.

\paragraph{Classification Model Distillation.}  
For classification model distillation, we use the \texttt{gte-Qwen2.5-1.5B-instruct}\footnote{https://huggingface.co/Alibaba-NLP/gte-Qwen2-1.5B-instruct}~\cite{li2023towards} checkpoint as the encoder backbone. The training dataset consists of synthetic job–person pairs annotated by the teacher model with fit labels, and is several orders of magnitude larger than the dataset used for training the in-house teacher model, allowing the classifier to generalize across a broader range of job–person fit scenarios.

The input is encoded using the LLM encoder, and the resulting embeddings are passed through a lightweight MLP head to produce classification distribution. We train the model for 10 epochs with a learning rate of $1 \times 10^{-4}$ and a maximum sequence length of 8192.

\subsection{Baselines and Evaluation Metrics}

\subsubsection{Explanation Task.}  
We evaluate \OURS on the explanation task using ROUGE-1, ROUGE-2, and ROUGE-L as the primary metrics~\cite{lin2004rouge}, comparing generated outputs against teacher references. We benchmark our method against standard supervised finetuning (SFT), where the model is trained directly on the teacher outputs without any knowledge distillation. In addition to ROUGE scores, we report generation loss to assess how well the student mimics the teacher’s output distribution.

\subsubsection{Classification Task.}  
For the classification task, we compare several model architectures, including (1) base structure (sequence classification or two tower) (2) embedding pooling method (last token or mean pooling) (3) interaction of job and profile embeddings followed by an MLP classifier. We report weighted F1 scores to evaluate classification performance across multiple fit levels (e.g., low, medium, high). \OURS consistently outperforms these baselines by leveraging teacher-guided supervision via distillation.

\section{Experiment Results}
In this section, we present the main results in the job-person fit tasks. 

\subsection{Main Results}
We present the main results in three parts. Section~\ref{sec:teacher-result} reports the performance of the teacher model. Section~\ref{sec:explanation-result} details the results for the explanation task. Section~\ref{sec:classification-result} presents the performance on the classification task.

\subsubsection{Teacher model finetuning results}\label{sec:teacher-result}

We present the teacher model finetuning results in Table~\ref{tab:teacher-results}. Following the training setup described in Section~\ref{sec:training-detail}, we finetune three models from the \texttt{Qwen} series: \texttt{Qwen3-8B}~\cite{yang2025qwen3}, \texttt{Qwen2.5-7B}, and \texttt{Qwen2.5-7B-instruct}. Among these, \texttt{Qwen2.5-7B-instruct} achieves the best performance across all ROUGE metrics, demonstrating its superior suitability as a teacher model under the same training configuration. Hence, we use this finetuned \texttt{Qwen2.5-7B-instruct} model in the following sections, as the inhouse teacher model.

\begin{table*}[t]
\centering
\resizebox{0.9\textwidth}{!}{
\begin{tabular}{l|l|l|c|ccc}
    \toprule
    \textbf{Path} & \textbf{Teacher Model} & \textbf{Student Model} & \textbf{Loss} & \textbf{ROUGE\_1} & \textbf{ROUGE\_2} &  \textbf{ROUGE\_L} \\
    \midrule
    Single-stage (7B → 0.5B) & Qwen2.5-7B & Qwen2.5-0.5B & 0.1274 & 0.7136 & 0.5736 & 0.6056 \\ 
    Single-stage (7B → 1.5B)& Qwen2.5-7B & Qwen2.5-1.5B & 0.1167 & 0.7820 & 0.6390 & 0.6715 \\ 
    \textbf{Single-stage (7B → 3B)} & \textbf{Qwen2.5-7B} & \textbf{Qwen2.5-3B}   & \textbf{0.1128} & \textbf{0.7917} & \textbf{0.6480} & \textbf{0.6798} \\ 
    \midrule
    2-stage (7B → 3B → 1.5B)      & Distilled-3B & Qwen2.5-1.5B & 0.1131 & 0.7769 & 0.6346 & 0.6659 \\ 
    2-stage (7B → 3B → 0.5B)      & Distilled-3B & Qwen2.5-0.5B & 0.1240 & 0.7721 & 0.6259 & 0.6594 \\
    2-stage (7B → 1.5B → 0.5B)      & Distilled-1.5B & 
    Qwen2.5-0.5B & 0.1234 & 0.7673 & 0.6233 & 0.6535 \\
    \bottomrule
\end{tabular}
}
\caption{Ablation results for different knowledge distillation paths. Multi-stage pipelines use intermediate student models as teachers for subsequent stages.}
\label{table:ablation-results}
\end{table*}

\begin{table}[h]
\centering
\resizebox{0.5\textwidth}{!}{
\begin{tabular}{l|ccccc}
    \toprule
    \textbf{Teacher Model} & \textbf{ROUGE\_1} & \textbf{ROUGE\_2} &  \textbf{ROUGE\_L}\\
    \midrule
    Qwen3-8B &  0.7918 & 0.6566 & 0.6902 \\
    Qwen2.5-7B &  0.7766  & 0.6328 & 0.6655 \\
    \textbf{Qwen2.5-7B-instruct} & \textbf{0.8084} & \textbf{0.6835} & \textbf{0.7156} \\
    \bottomrule
\end{tabular}
}
\caption{Teacher model finetuning results.}
\label{tab:teacher-results}
\end{table}

\subsubsection{Explanation task results}\label{sec:explanation-result}

We evaluate the performance of various knowledge distillation methods on the explanation task, using \texttt{Qwen2.5-7B-instruct} as the teacher model and \texttt{Qwen2.5-1.5B} series as the student model. As shwon in Table~\ref{table:main-results}, all KD variants yield substantial improvements in ROUGE scores, indicating effective knowledge transfer learned in the training process.
Among the four mentioned KD methods, TVD achieves the highest ROUGE scores, suggesting that it facilitates better alignment with the teacher outputs, while maintaining stable training dynamics.
Although FKL results in the lowest total loss, its ROUGE scores are slightly lower than those of TVD, highlighting a trade-off between minimizing the distillation loss and maximizing generation quality. SKL and JSD also perform competitively, with JSD achieving lowest KD loss overall.
The above results demonstrate that, using token-level KD objectives is effective, and suggest that TVD strikes the best balance between alignment and task performance in the explanation task settings.

For the final explanation model candidates, we conducted human evaluations as described in Section~\ref{sec:human-eval}. Results show that the distilled model performs on par with—or in some cases slightly better than—the proprietary teacher model. We attribute this to two factors: (1) the evaluation setting focuses on a narrow domain with well-defined and repetitive patterns, making it easier for the student model to generalize; and (2) the training data for the student model was extensively curated, including filtering steps that removed low-quality or inconsistent outputs from the teacher.

\subsubsection{Classification task results}\label{sec:classification-result}
We present the classification task results in Table~\ref{tab:classification-results-relative}. The model performance is evaluated under different model structure and pooling configs, as summarized in Table~\ref{tab:classification-results-relative}. Two backbone model series are compared, Phi3-mini and gte-Qwen2.5-1.5B-instruct, across two model structures: sequence classification (SeqCls) and two-tower(TwoTower) models. Within each structure, we further explore different pooling and interaction strategies. 
As shown in the result table, models from \textbf{gte-Qwen2.5-1.5B-instruct} consistently outperform those from Phi3-mini\footnote{https://huggingface.co/collections/microsoft/phi-3-6626e15e9585a200d2d761e3}~\cite{abdin2024phi3technicalreporthighly}, across both accuracy and F1-score.
The best-performing configuration is the SeqCls model with last-token pooling, achieving an accuracy improvement of +1.39\% and an F1-score of +0.42\%. This suggests that the last token representation in the encoder captures sufficient semantic information for fit classification when trained with supervision.
However, two-tower models while more modular and potentially more scalable for retrieval tasks, generally underperform compared to the SeqCls counterparts. For example, in gte-Qwen2.5-1.5B-instruct models, two-tower models with encoding with a unified sequence may better capture the nuanced interplay between job and member features in this setting.


\begin{table}[h]
\centering
\resizebox{0.5\textwidth}{!}{
\begin{tabular}{l|ccccc}
    \toprule
    \textbf{Backbone Model} & \textbf{Structure} & \textbf{Pooling} & \textbf{Interaction} & \textbf{Accuracy} & \textbf{F1-score} \\
    \midrule
    \multirow{4}{*}{Phi3-mini} 
        & \multirow{2}{*}{SeqCls} 
            & Last & - & -- & -- \\
        &       & Mean & - & -0.08\% & -0.22\% \\
        & \multirow{2}{*}{TwoTower} 
            & Last & Concat & -1.69\% & -2.21\% \\
        &       & Last & DotProduct & -1.58\% & -2.02\% \\
    \midrule
    \multirow{4}{*}{\textbf{gte-Qwen2.5-1.5B-instruct}} 
        & \multirow{2}{*}{\textbf{SeqCls}} 
            & \textbf{Last} & \textbf{-} & \textbf{+1.39\%} & \textbf{+0.42\%} \\
        &       & Mean & - & +1.26\% & +0.18\% \\
        & \multirow{2}{*}{TwoTower} 
            & Last & Concat & +1.03\% & -3.25\% \\
        &       & Last & DotProduct & +0.55\% & -3.61\% \\
    \bottomrule
\end{tabular}
}
\caption{Relative performance on the classification task. Phi3-mini with SeqCls and Last pooling is used as the baseline. Other values represent absolute changes in accuracy and F1-score. Due to confidential reasons, we cannot disclose absolute numbers here.}
\label{tab:classification-results-relative}
\end{table}

\subsection{Analysis: Multi-Stage Distillation Paths}

To assess the effectiveness of multi-stage knowledge distillation, we compare single-stage and two-stage strategies for compressing a 7B teacher model into a 0.5B student model on the explanation generation task, as shown in Table~\ref{table:ablation-results}. The Distilled-3B, Distilled-1.5B, and Distilled-0.5B models denote student models distilled directly from the Qwen2.5-7B teacher. The single-stage approach directly distills the 0.5B student from the 7B teacher, achieving a ROUGE-1 score of 0.7136.

In contrast, the two-stage setting introduces an intermediate teacher to bridge the capacity gap. For instance, in the 7B $\rightarrow$ 1.5B $\rightarrow$ 0.5B pipeline, the 7B teacher is first distilled into a 1.5B model, which then serves as the teacher for the 0.5B student. This approach achieves a ROUGE-1 score of 0.7726, representing an absolute improvement of +0.059. A similar two-stage path using a 3B intermediate (7B $\rightarrow$ 3B $\rightarrow$ 0.5B) also yields substantial gains (ROUGE-1 = 0.7721), suggesting that intermediate teachers of different capacities can both facilitate knowledge transfer effectively. ROUGE-2 and ROUGE-L scores show consistent improvements, indicating better preservation of both content and fluency.

These results suggest that direct compression from a large teacher to a small student may suffer from representational mismatch and optimization challenges. Incorporating an intermediate model alleviates these issues by enabling a smoother transfer of knowledge. While two-stage distillation incurs additional training cost, it significantly enhances explanation quality and is particularly beneficial in low-capacity regimes.

\section{Online Deployment}
We deploy the LANTERN models in production on NVIDIA A100 GPUs to serve real-time classification and explanation requests. The classification model powers job–person fit levels shown directly on job description page, while the explanation model is triggered on demand when users request detailed fit explanation. The system is optimized to balance throughput and latency under peak load. We also present key performance metrics and implementation strategies that improve scalability and efficiency.

\subsection{System Optimizations}

As discussed in Section~\ref{sec:serving}, we apply input compression to job descriptions to reduce serving cost. We tested different compression rates and found that reducing input length by about 80\% preserves key requirement information without affecting downstream performance

Given this reduced input, we further downscale the classification model from fine-tuning a \texttt{gte-Qwen2.5-1.5B-instruct} model to a \texttt{ModernBERT}~\cite{warner2024smarter} variant with 0.4B parameters, which achieves comparable offline performance with negligible degradation, leading to a 4$\times$ improvement in serving throughput per GPU.

To support multilingual input scenarios, we keep the 1.5B model as a fallback option. However, it is invoked selectively and does not impact the primary serving path for English-dominant workloads.

\subsection{Throughput and Latency}



Table~\ref{table:serving-pipeline} summarizes the peak traffic, throughput (QPS per GPU), and latency for each component in the LANTERN serving pipeline. To ensure efficient real-time serving, we prioritize throughput provisioning based on the observed traffic volume while ensuring latency remains within product constraints.

Traffic and throughput values are normalized to the explanation module, which receives the lowest number of requests, as it is only triggered when a user clicks to view detailed match rationale. In contrast, the classification module handles the highest request volume, powering fit levels shown immediately in job search and browsing. The summarization module is triggered less frequently, typically once per job posting creation or update.

These varying usage patterns lead to imbalanced serving loads: the classification module receives approximately 30$\times$ more requests than explanation, while summarization receives 4$\times$ more. To accommodate this imbalance, we provision serving capacity based on traffic volume, scaling the classification model to a lightweight 0.4B encoder that achieves 80$\times$ the throughput of the explanation model while maintaining low latency (P95 of 0.3s). In contrast, summarization and explanation rely on larger decoder models and operate at lower throughput, but still meet product latency requirements (P95 of 3.5s and 7.2s, respectively).


\begin{table}[htbp]
\centering
\resizebox{0.5\textwidth}{!}{
\begin{tabular}{l|c|c|c|c|c}
    \toprule
    \textbf{Module} & \textbf{Type} & \textbf{Size} & \textbf{Peak Traffic} & \textbf{Throughput} & \textbf{Latency(P95)} \\
    \midrule
    Summarization & Decoder & 1.5B & 4$\times$ & 4$\times$ & 3.5s \\
    Classification & Encoder & 0.4B & 30$\times$ & 80$\times$ & 0.3s \\
    Explanation & Decoder & 1.5B & 1$\times$ & 1$\times$ & 7.2s \\
    \bottomrule
\end{tabular}
}
\caption{Peak traffic, throughput (QPS per GPU), and latency for each component.}
\label{table:serving-pipeline}
\end{table}

\section{Conclusion}
In this paper, we introduced LANTERN, a scalable knowledge distillation framework for job–person fit and explanation. We shared practical techniques for synthetic data generation, prompt engineering, and multi-level distillation to build lightweight classification and explanation models from a powerful black-box teacher and an in-house white-box teacher. The framework addresses key challenges in latency, scalability, and output quality for domain-specific LLM applications. LANTERN has been deployed in production at LinkedIn, powering job–person evaluation with measurable gains in apply rate and qualified applications. We believe the insights in this paper can benefit industry practitioners building efficient and interpretable LLM-based systems at scale.

\bibliographystyle{ACM-Reference-Format}
\bibliography{sample-base}

\end{document}